\documentclass[journal,twoside]{IEEEtran}
\pdfoutput=1

\usepackage{amsmath, amssymb, amsfonts, amsthm}
\usepackage{thmtools}
\usepackage{empheq}
\usepackage{siunitx}
\theoremstyle{definition}

\usepackage[switch, pagewise]{lineno}
\usepackage[colorlinks]{hyperref}
\usepackage[noabbrev]{cleveref}

\crefname{equation}{Eq.}{Eqs.}
\crefname{figure}{Fig.}{Figs.}
\crefname{algorithm}{Algorithm}{Algorithms}
\crefname{table}{Table.}{Tables.}
\Crefname{table}{Table.}{Tables.}
\crefname{table*}{Table.}{Tables.}
\Crefname{table*}{Table.}{Tables.}

\crefname{lem}{Lemma.}{Lemmas.}
\Crefname{lem}{Lemma.}{Lemmas.}
\crefname{thm}{Theorem.}{Theorems.}
\Crefname{thm}{Theorem.}{Theorems.}
\crefname{prop}{Proposition.}{Propositions.}
\Crefname{prop}{Proposition.}{Propositions.}

\usepackage{algorithm}
\usepackage{algorithmic}

\usepackage{subfloat}
\usepackage{graphicx}
\usepackage[caption=false, font=footnotesize]{subfig}

\usepackage{filecontents}
\usepackage{pgfplots}
\usepackage{tikzscale}
\usepackage{tikz}
\usepgfplotslibrary{patchplots}

\usepackage{multirow}
\usepackage{makecell}

\title{Knowledge Distillation for Oriented Object Detection on Aerial Images}

\author{
		Yicheng Xiao,
		Junpeng Zhang, ~\IEEEmembership{Member, ~IEEE,} }
		 
\begin{document}
\pgfplotsset{compat=1.14}
	
\maketitle

\begin{abstract}
	
Deep convolutional neural network with increased number of parameters has achieved improved precision in task of object detection on natural images, where objects of interests are annotated with horizontal boundary boxes. 
On aerial images captured from the bird-view perspective, these improvements on model architecture and deeper convolutional layers can also boost the performance on oriented object detection task. 
However, it is hard to directly apply those state-of-the-art object detectors on the devices with limited computation resources, which necessitates lightweight models through model compression. 
In order to address this issue, we present a model compression method for rotated object detection on aerial images by knowledge distillation, namely KD-RNet. 
With a well-trained teacher oriented object detector with a large number of parameters, the obtained object category and location information are both transferred to a compact student network in KD-RNet by collaborative training strategy. 
Transferring the category information is achieved by knowledge distillation on predicted probability distribution, and a soft regression loss is adopted for handling displacement in location information transfer. 
The experimental result on a large-scale aerial object detection dataset (DOTA) demonstrates that the proposed KD-RNet model can achieve improved mean-average precision (mAP) with reduced number of parameters, at the same time, KD-RNet boost the performance on providing high quality detections with higher overlap with groundtruth annotations.
	
\end{abstract}


\IEEEpeerreviewmaketitle

\section{Introduction}

\begin{figure}
    \centering
    \includegraphics[width=0.925 \linewidth]{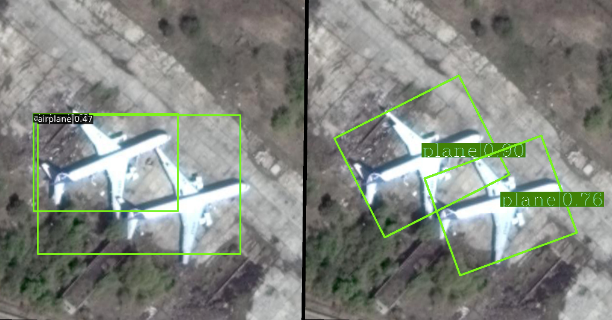}
    \caption{Comparison on the horizontal annotation and rotated annotation for oriented objects. }
    \label{fig:comp_hbb_obb}
\end{figure}

\IEEEPARstart{I}{n} recent years, object detection task on natural images has observed boosted detection precision with a variety of deep convolutional networks sharing different model architectures, such as Faster-RCNN \cite{r16}, SSD \cite{r17}, RetinaNet \cite{r1}, FCOS \cite{r6} and etc. 
Most of these models are originally designed for the objects with horizontal boundary boxes.
For images captured from bird-view, such as aerial images, objects of interests are commonly of arbitrary orientations, and increased background areas are included within a horizontal boundary box, as demonstrated \Cref{fig:comp_hbb_obb}.
In order to handle the object orientation on aerial images, a variety of angle-based and polygon boundary box encoding are developed \cite{r7,r8,r9,r10,r11,r12,r13}, and strategy for extracting rotated region proposals are proposed for further boosting the detection precision \cite{r7}. 
However, these detectors are commonly over-parameterized, and they can hardly be deployed on the devices with limited computation resources for onboard processing. 

In order to alleviate the burden of computation consumption of large network models, compact models can be obtained by pruning, quantification \cite{r14}, weight sharing, transfer/compact convolution filters \cite{r4}. 
Alternatively, model compression can also be formulated as a knowledge distillation problem, where distribution predicted by a large teacher network is transferred to a small student network through knowledge distillation \cite{r15,r5,r2,r3}. 
To obtain compact object detector with comparable detection precision, different knowledge distillation strategies are proposed for horizontal object detection tasks. 
However, for rotated object detection in remote sensing images, model compress using knowledge distillation technique has not been well explored.

To handle object orientation in knowledge distillation, we present a strategy to mimic the behaviors of both classification and boundary box regression branches from a teacher detector during training a compact student model, namely KD-RNet. 
Inspired by the soft targets in distilling category knowledge in classification task, the semantic knowledge on object category from a teacher model is transferred through soft-target-based knowledge distillation.
Location displacement predicted by the teacher model is another type of information for guiding a compact student model, and we propose a soft regression loss for transferring such location knowledge.
We validate the effectiveness of the proposed KD-RNet by transferring a single-stage object detector to a compact student model with a shadow backbone, and the experiments are conducted on a large-scale aerial object detection dataset (DOTA).
The experimental results demonstrates that the proposed KD-RNet model can achieve improved mean-average precision (mAP) with reduced number of parameters, at the same time, KD-RNet is also able to boost the performance on providing high quality detections with higher overlap with groundtruth annotations.

\begin{figure*}[t]
    \centering
    \includegraphics[width=0.925 \linewidth]{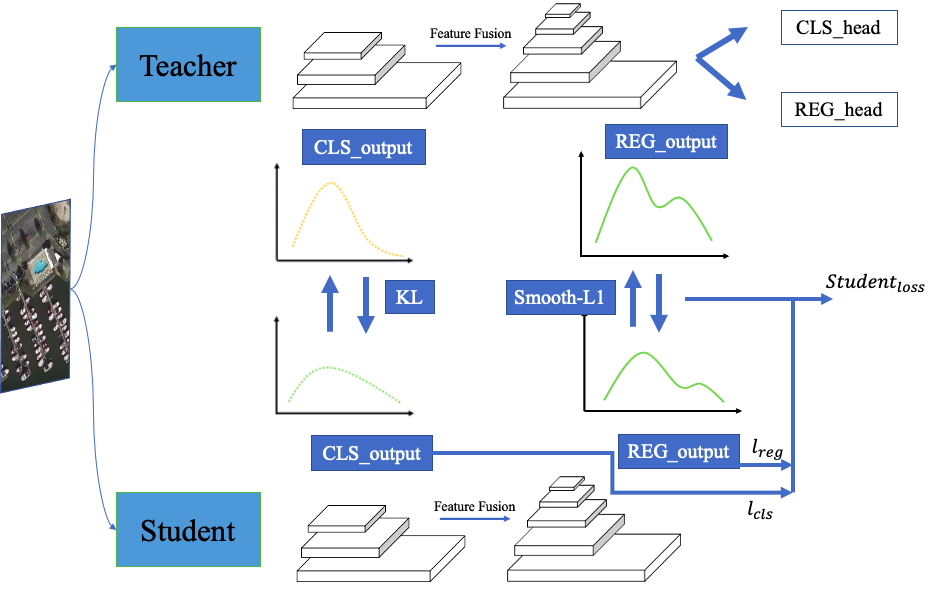}
    \caption{The framework of the proposed KD-RNet model.}
    \label{fig:framework}
\end{figure*}

The remainder of this paper is organized as follows. 
The proposed KD-RNet model is presented in \Cref{sec:method}, which is followed by the experimental results reported in \Cref{sec:experiments}.
Finally, conclusions and suggestions for future research are given in Section \ref{sec:conclusion}.

\section{KD-RNet}
\label{sec:method}

\subsection{Framework}

KD-RNet aims to transfer the semantic knowledge and location knowledge obtained from a teacher detector to a compact student model. 
As shown in \Cref{fig:framework}, KD-RNet is compose of a well-trained teacher detector and a compact student model, and knowledge from the teacher model is employed for guiding the training of the student model.
During this procedure, we pay attention to not only the classification information of anchor but also the location information, which is achieved by taking the predicts from the teacher's detection head as soft learning targets for student models, and combine such soft targets with hard target provided by groundtruth annotation in calculating the loss for back-propagation optimization of the student model.
In this report, we choose a classical single-stage network, RetinaNet, as a experienced teacher, and a similar network with a smaller backbone is adopted as a compact student model.

\subsection{Knowledge Distillation Process}

Updating the parameters of the student model in KD-RNet is driven by the loss between model predictions, soft-targets from the teacher model and hard-targets from the groundtruth annotation, and the overall loss function of the student network can be expressed as
\begin{equation}
    \mathcal{L}_{all} = \lambda_{1} \mathcal{L}_{cls} + \lambda_{2} \mathcal{L}_{reg} +\lambda_{3} \mathcal{L}_{KD-cls} + \lambda_{4} \mathcal{L}_{KD-reg},
\end{equation}
where $\mathcal{L}_{cls}$ and $\mathcal{L}_{reg}$ refer to the classification and regression losses between the student model's prediction and groundtruth, respectively.
$\mathcal{L}_{KD-cls}$ and $\mathcal{L}_{KD-reg}$ denote the distillation losses on anchor category and object location, and the scalars $\lambda_{i}$, $\forall i \in \{1 ,2, 3, 4\}$, are used for balancing the contribution of different items to the overall loss $\mathcal{L}_{all}$.

For the classification branch of the student model, the we deploy the relative KL divergence as the distillation loss on anchor category $\mathcal{L}_{KD-cls}$ by describing the distance between the predicted distributions by the teacher model and the student model in KD-RNet. 
Let $P_{s}$ denote as the probabilities of classification prediction of student model, and $P_{t}$ refers to the ones from the teacher model.
The distillation loss on anchor category $\mathcal{L}_{KD-cls}$ can be formulated as 
\begin{equation}
    \mathcal{L}_{KD-cls} = -\sum_{x \in X} p_{s}(x) \log(p_t(x)) + \sum_{x \in X} p_{s}(x) \log(p_s(x)).
\end{equation}
In term of the classification loss with hard target $\mathcal{L}_{cls}$, we adopt the Focal Loss \cite{r1} between model predicts and anchor labels assigned by groundtruth.

\begin{table*}[h]
	\caption{Ablation study on using different knowledge distillation strategies}
	\label{tbl:ab_study}
	\centering
	\begin{tabular}{c|c|c|c|c|c|c}
	    \hline
        Network &   Backbone    &	KD-cls  &	KD-reg   &	mAP(DOTA) & AP$_{75}$  & Param  \\  
        \hline
        RetinaNet    &   ResNet50    &	-   &	-	&   71.47   &   37.44   &	292.6M \\
        \hline
        RetinaNet    &   ResNet34	 &  -   &	-	&   70.20	&   34.08   &   240.8M\\
        \hline
        KD-RNet     &   ResNet34    &  	KL	&   Smooth-L1   &	70.45   &   36.70   &	240.8M  \\
        \hline
        KD-RNet     &	ResNet34    &   KL	&   -	&   70.20   &   -   &	240.8M \\
        \hline
        KD-RNet     &   ResNet18    &	KL  &	Smooth-L1   &	68.27   &   35.85&	160.2M\\
		\hline
	\end{tabular} 
\end{table*}

For the regression task, we adopted KLD loss \cite{r18} for measuring the regression loss $\mathcal{L}_{reg}$ between predicts bounding box and ground truth box.
At the same time, the predicted location displacement by the teacher model is also used for guiding the student model by a soft regression loss
\begin{equation}
    \mathcal{L}_{KD-reg}=
    \begin{cases}
    \frac{1}{2}(x_{t} - x_{s})^2, & |x_t - x_s |\leq 1 \\
    |x_{s} - x_{t}| - \frac{1}{2}, & \text{others}
    \end{cases},
\end{equation}
where $x_t$ and $x_s$ refer to the regression predictions by teacher model and student model, respectively.
By combining soft and hard losses, KD-RNet is able to transfer the obtained category and location information from a experienced teacher model to a compact student model, which is experimentally validated in the following section.

\section{Experiments}
\label{sec:experiments}

\subsection{Dataset and Implementation Details}

DOTA \cite{r19} is a large-scale aerial object detection benchmark that contains 2,806 aerial images from different sensors and platforms. 
The image size ranges from around $800 \times 800$ to $4, 000 \times 4, 000$ pixels with objects exhibiting a wide variety of scales, orientations, and shapes. 
These images are then annotated by experts using 15 common object categories. 
The fully annotated DOTA benchmark contains 188,282 instances labeled with rotated boundary boxes. 
We follow the official guideline for splitting the datasets -- half of the original images as the training set, $1/2$ as the validation set, and $1/3$ as the testing set. 
For each image, we extract a set of $600 \times 600$ subimages with an overlap of 150 pixels and 300 pixels for train set and test test, respectively.

In KD-RNet, In the classification branch, we set $\gamma = 2.0$ and $\alpha = 0.25$ in Focal Loss as usual. 
Empirically the temperature in the distillation loss of classification branch and regression branch are set as 8 and 10, respectively.

\subsection{Performance Comparison}

By introducing knowledge distillation in training a compact student detector, our KD-RNet is more likely to ease the training procedure and achieve improved detection precision, especially for high quality detections.
As shown in \Cref{tbl:comp_res_backbone}, KD-RNet outperforms that the same network trained from scratch by $1.03\%$ in the term of mAP(COCO).
At the same time, we observed that using knowledge distillation on regression branch also works as a regularization term for preventing over-fitting, which can be validated by the increased precision during test by KD-RNet in \Cref{tbl:reg_term_effect}.

\begin{table}[h]
	\caption{Performance comparison with models using backbones in difference scales}
	\label{tbl:comp_res_backbone}
	\centering
	\begin{tabular}{c|c|c|c|c}
		\hline
		Network &   BackBone    & mAP (DOTA)    &   AP$_{75}$ & mAP (COCO) \\
		\hline
		RetinaNet   &   ResNet50 &  71.47   &   37.44   &   39.06   \\
		\hline
		
		RetinaNet   &   ResNet34 &  70.20   &   34.08   &   37.51   \\
		KD-RNet     &   ResNet34 &  70.45   &   36.70   &   38.54   \\
		\hline 
		
		RetinaNet   &   ResNet18 &  68.37   &   35.68   &   37.61   \\
		KD-RNet     &   ResNet18 &  68.27   &   35.85   &   37.32   \\
		\hline
	\end{tabular} 
\end{table}

\begin{table}[h]
	\caption{Performance on training and test sets}
	\label{tbl:reg_term_effect}
	\centering
	\begin{tabular}{c|c|c|c}
	\hline
        Network &   Backbone &	mAP(test)   &	mAP(train) \\
        \hline
        RetinaNet   & ResNet34  &	70.20   &	82.54 \\
        \hline
        KD-RNet  & ResNet34  &	70.45   &	81.75 \\
		\hline
	\end{tabular} 
\end{table} 




\section{Conclusion}
\label{sec:conclusion}

We apply knowledge distillation method to object detector to make it perform well in remote sensing field. 
In the stage of knowledge distillation, we not only focus on the classification information of teacher network, but also consider that the location information of object bounding box is worth learning. 
Therefore, in the distillation process, we also learn the information of regression network by our proposed method.
The results show that our distillation method can improve the accuracy of the model with the same number of parameters, while for the small student model which is with the smaller numbers of parameters, the performance is good enough especially for high precision prediction.

\bibliographystyle{IEEEtran}
\footnotesize\bibliography{main}

\end{document}